\documentclass[conference]{IEEEtran}
\IEEEoverridecommandlockouts
\usepackage{wrapfig}
\usepackage{tabularx}
\usepackage{multirow}
\usepackage{cite}
\usepackage{amsmath,amssymb,amsfonts}
\usepackage{algorithmic}
\usepackage{graphicx}
\usepackage{textcomp}
\usepackage{xcolor}
\usepackage{booktabs}
\usepackage{hyperref}
\usepackage{comment}
\def\BibTeX{{\rm B\kern-.05em{\sc i\kern-.025em b}\kern-.08em
    T\kern-.1667em\lower.7ex\hbox{E}\kern-.125emX}}
\begin{document}

\title{Electrical Grid Anomaly Detection \\via Tensor Decomposition\\
}

\author{\IEEEauthorblockN{1\textsuperscript{st} Alexander B. Most}
\IEEEauthorblockN{2\textsuperscript{nd} Maksim E. Eren}
\IEEEauthorblockA{\textit{Advanced Research in Cyber Systems} \\
\textit{Los Alamos National Laboratory}\\
Los Alamos, USA \\
AMost@lanl.gov\\
maksim@lanl.gov}

\and
\IEEEauthorblockN{3\textsuperscript{rd} Boian S. Alexandrov}
\IEEEauthorblockA{\textit{Theoretical Division} \\
\textit{Los Alamos National Laboratory}\\
Los Alamos, USA \\
}

\and
\IEEEauthorblockN{4\textsuperscript{th} Nigel Lawrence}
\IEEEauthorblockA{\textit{Advanced Research in Cyber Systems} \\
\textit{Los Alamos National Laboratory}\\
Los Alamos, USA
}
}

\maketitle

\begin{abstract}
Supervisory Control and Data Acquisition (SCADA) systems often serve as the nervous system for substations within power grids. These systems facilitate real-time monitoring, data acquisition, control of equipment, and ensure smooth and efficient operation of the substation and its connected devices. As the dependence on these SCADA systems grows, so does the risk of potential malicious intrusions that could lead to significant outages or even permanent damage to the grid. Previous work has shown that dimensionality reduction-based approaches, such as Principal Component Analysis (PCA), can be used for accurate identification of anomalies in SCADA systems. While not specifically applied to SCADA, non-negative matrix factorization (NMF) has shown strong results at detecting anomalies in wireless sensor networks. These unsupervised approaches model the normal or expected behavior and detect the unseen types of attacks or anomalies by identifying the events that deviate from the expected behavior. These approaches; however, do not model the complex and multi-dimensional interactions that are naturally present in SCADA systems. Differently, non-negative tensor decomposition is a powerful unsupervised machine learning (ML) method that can model the complex and multi-faceted activity details of SCADA events. In this work, we novelly apply the tensor decomposition method Canonical Polyadic Alternating Poisson Regression (CP-APR) with a probabilistic framework, which has previously shown state-of-the-art anomaly detection results on cyber network data, to identify anomalies in SCADA systems. We showcase that the use of statistical behavior analysis of SCADA communication with tensor decomposition improves the specificity and accuracy of identifying anomalies in electrical grid systems. In our experiments, we model real-world SCADA system data collected from the electrical grid operated by Los Alamos National Laboratory (LANL) which provides transmission and distribution service through a partnership with Los Alamos County, and detect synthetically generated anomalies.
\end{abstract}

\section{Introduction}
\begin{figure}[h]
    \centering
    \includegraphics[width=1\linewidth]{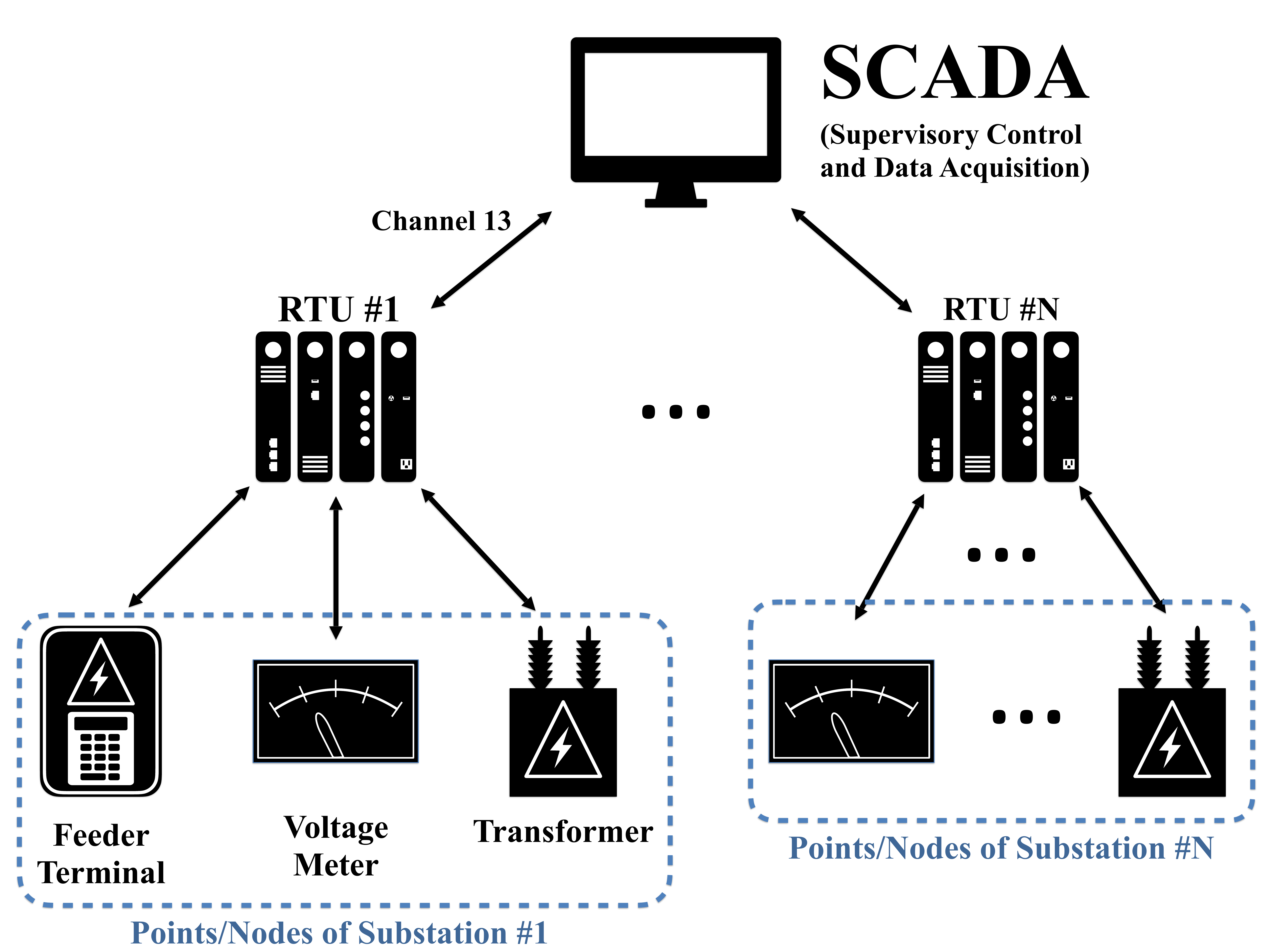}
    \caption{Diagram of an Electrical Grid SCADA System: Each substation is equipped with one Remote Terminal Unit (RTU) that gathers measurements from various devices (points) such as Feeder Terminals and Voltage Meters. The SCADA control system regularly sends requests to these RTUs, which in turn respond with their collected measurements.}
    \label{fig:scada}
\end{figure}
Power outages and interruptions within critical infrastructure represent a substantial threat to modern societies, impacting public safety, healthcare, and economic stability \cite{skarha2021poweroutage}. In 2008, the U.S. Department of Energy estimated that power outages cost Americans \$150 Billion annually \cite{Litos2008}. Illustrative examples of the devastating impact of critical infrastructure outages include the malware attacks on electrical grid operators in Ukraine in 2015 and 2016 that left hundreds of thousands of customers in the dark \cite{zetter2016ukraine, bbc2017ukrainepower}. The detection of such attacks and anomalous events in electrical grids can help in preventing such catastrophic outcomes. Therefore, there is an urgent need to improve and introduce anomaly detection methods to drive progress in the safety and security of critical infrastructure. 

Supervisory Control and Data Acquisition (SCADA) systems, which are pivotal in controlling and monitoring power grid operations, have become increasingly complex and vulnerable to both cyber and physical attacks. SCADA systems communicate using specific messages sent to and from Remote Terminal Units (RTUs) or Programmable Logic Controllers (PLCs) within a substation, allowing operators to remotely monitor equipment statuses, control circuit breakers, and gather telemetry data, among other tasks \cite{YADAV2021100433}. Given the critical nature of substations in managing and distributing electrical power, ensuring the integrity and reliability of electrical grid SCADA communication is paramount. An adversary with full access to an electrical grid SCADA system could potentially cause power outages and in some cases permanent damage to devices that make up the grid. Lloyds and the University of Cambridge's Center for Risk Studies estimate that a coordinated large scale attack on sections of the U.S. power grid could cause \$243 billion to \$1 trillion in damage \cite{Lloyds}. Early detection is key to mitigating these attacks, and anomaly detection systems provide a key mechanism to rapidly identify malicious behavior. Electrical grid SCADA systems transmit messages at a very high frequency. Because the connectivity and complexity of these systems is constantly increasing, their transmission rate will only increase in the future. Detecting anomalies amidst this flood of messages in a constantly evolving network requires sophisticated analysis techniques. 

Traditional anomaly detection techniques might struggle to capture the nuances of SCADA system messages. Previous work has shown that dimensionality reduction techniques such as Principal Component Analysis (PCA) \cite{doi:10.1080/14786440109462720, 9042509, 7573322} and Non-negative Matrix Factorization (NMF) \cite{lee1999learning, GraphLink2020} learn accurate predictive models of computer networks \cite{8450421}. However, these methods do not model the complex higher-dimensional details of SCADA systems. To this end, while PCA and NMF demonstrated some success in detecting anomalies generated by less sophisticated adversaries, they quickly falter as intrusions become more sophisticated and more closely resemble normal behavior.

In response to these challenges, this paper applies Canonical Polyadic Alternating Poisson Regression (CP-APR), a powerful non-negative tensor factorization method, for modeling multi-faceted details hidden in data and detecting anomalies in SCADA systems. Specifically, we use the publicly available version of CP-APR, named pyCP-APR, with statistical anomaly detection framework \cite{Eren2021pyCPAPR, 10.1145/3519602}. \cite{10.1145/3519602} demonstrated leading-edge anomaly detection results on cyber network data using CP-APR. Here we selected to use CP-APR given its ability to efficiently factorize sparse tensor representations. In our experiments, CP-APR is trained on real historical SCADA messages and tested on synthetic anomalies. CP-APR shows an improvement over our baselines PCA and NMF when detecting a wide range of power grid attacks. By capturing the unique patterns across all dimensions in SCADA messages, this research offers a robust and more effective method for enhancing security within these vital networks. To the best of our knowledge, we are the first to apply a tensor decomposition based anomaly detection, via statistical framework, for SCADA system security. Our contributions include:

\begin{itemize}
    \item Applying non-negative tensor decomposition to SCADA system anomaly detection under a statistical framework.
    \item Comparing of anomaly detection via PCA and NMF to the higher-dimensional anomaly detection with tensor decomposition. 
    \item Showcasing the SCADA system anomaly detection capabilities using several distinct configurations of tensors built from SCADA communication data.
    \item Addressing the critical infrastructure security problem with data collected from real-world SCADA system at Los Alamos National Laboratory, complimented with synthetic anomalies. 
\end{itemize}

\section{Background}
 
SCADA system anomalies might indicate cyber-attacks or system malfunctions. This section summarizes prior methods for electrical grid anomaly detection. 

Several neural networks have been proposed for learning normal SCADA behavior to detect anomalies. \cite{5706699} developed a three-stage supervised neural network for intrusion detection of messages in a water treatment plant SCADA network. Despite the accuracy of deep learning-based solutions, the need for labeled training data, often scarce or costly, remains a drawback. The diversity of SCADA systems makes it particularly difficult to acquire representative training data that would enable the broad application of supervised approaches. \cite{10.1145/3264888.3264896} treated measurements collected in a SCADA system as an unlabeled time series to learn normal behavior and showed that a 1d CNN could outperform an LSTM for anomaly detection. In comparison, we use an unsupervised method on the values present in outgoing SCADA messages, where we can model expected behavior without the need of specific attack signatures. 

\cite{sontowski2022detecting} used dynamic time warping with a SCADA historian and an automatic meter reading system to find overlapping measurement points for a similar dataset, and then used autoregression, rolling average and level shift to detect anomalies in time series of device measurements in the grid. While that effort focused on divergence in measurement trends between independent measurement systems, it did not examine the underlying messages directly. Because SCADA messages themselves provide additional information not present in the raw measurements, they can be used to identify behavior that is undetectable via measurements alone. 

Several forms of Matrix factorization have been used for anomaly detection in network data in the past. Principal Component Analysis (PCA) and Non-negative Matrix Factorization (NMF) have been popular due to their speed and performance on black-box attacks \cite{7573322,8450421}. However, our results show that these methods lose effectiveness when an adversary has more knowledge of the SCADA system, or when the attacks resemble the normal or expected behaviour more closely. \cite{Marton_Sanchez_Carlos_Martorell_2013} used PCA to select features for a partial least squares regression to detect abnormal behaviors in industrial equipment. \cite{7573322} used PCA to detect anomalies in primary distribution voltage magnitude measurements across an electrical grid SCADA network. PCA is a dimensionality reduction technique, but patterns in SCADA messages can exist across multiple dimensions that can be better modeled with a tensor decomposition approach. PCA represents the dataset in matrix form, and decomposes the matrix into K orthonormal vectors to describe the highest variance in the matrix. This orthogonality constraint can be problematic when patterns do not have orthogonal solutions.   

While we were unable to identify examples of NMF being used specifically on electrical SCADA systems, NMF has been used as an anomaly detection technique on similar problems and is closely related to tensor decomposition \cite{10.1145/3624567}. \cite{1428613} and \cite{8450421} used NMF to detect anomalies in computer network data and wireless sensor networks. \cite{TurcotteMHM16} and \cite{GraphLink2020} advance NMF for anomaly detection with Poisson Matrix Factorization (PMF). PMF is tailored for sparse count data modeled using Poisson distributions. We utilize Poisson Tensor Decomposition, an extension of PMF, to capture non-orthogonal patterns more accurately in tensor form than matrix form.

 While \cite{Ajayan2017ASO} demonstrated the abilities of Tucker Tensor Decomposition at visualizing patterns in SCADA measurements, to the best of our knowledge, anomaly detection using Tensor Decomposition has never been demonstrated on SCADA system messages. Additionally, \cite{sandoval2020threeway} used tensor decomposition on wattage data across Texas to detect anomalous power usage, which could occur from theft, illicit activities, or faulty devices. 

\section{Multidimensional Anomaly Detection}

We utilized the tensor anomaly detection framework, pyCP-APR, from our 2021 R\&D 100 winning SmartTensors AI Platform \cite{Eren2021pyCPAPR, SmartTensors, 10.1145/3519602}. pyCP-APR is a publicly available Python library based on CP-APR \cite{Chi2011, Eren2021pyCPAPR}. CP-APR was originally introduced by \cite{Chi2011} in 2010, then repurposed by \cite{Eren2020, 10.1145/3519602} for cyber network anomaly detection. For completeness, we will briefly summarize tensor decomposition and CP-APR here following the prior work \cite{10.1145/3519602}.

A tensor is a multidimensional array. SCADA system messages can be represented as a tensor with dimensions \textit{RTU ID, Number of Points Requested, Channel, and Time Since the Last Message to the Same RTU}. Here, the number of times a message was sent to an RTU with the same number of points requested, channel and periodicity would be the count value at the index of those dimension values. Tensor decomposition is a process of breaking down a tensor into its constituent parts, typically in a manner that exposes some underlying structures in the data. This process is analogous to matrix factorization for higher dimensions. Unlike matrix-based methods which often lose multidimensional relationships due to dimension reduction, tensor decomposition is not bound by orthogonality. Tensor decomposition can represent the original tensor using fewer components, thereby reducing noise, revealing patterns, saving storage and computation, and making the data more interpretable. 

Canonical Polyadic Decomposition (CPD), or CANDECOMP/PARAFAC, is a popular method for tensor decomposition. CPD breaks down a tensor into a sum of rank-one tensors. The Rank of CPD is the minimum number of rank-one tensors whose sum can reconstruct the original tensor. A rank-one tensor is a tensor that can be written as the outer product of $D$ vectors, where $D$ is the number of dimensions. For a tensor $\mathcal{X}$ where $D=3$, it would be represented as:

\begin{equation}
\mathcal{X} \approx \sum_{r=1}^{R} a_r \circ b_r \circ c_r
\end{equation}
Where \( \circ \) denotes the outer product, and \(a_r\), \(b_r\), and \(c_r\) are vectors. \( R \) is the rank of the decomposition.

Chi and Kolda found in \cite{Chi2011} that determining factors by negative log-likelihood of Poisson distributions could better describe the zeros in sparse count data when compared to the standard Gaussian assumption used in CPD. This work introduced a new variation of CPD called Canonical Polyadic Alternating Poisson Regression (CP-APR).  

CP-APR uses a multiplicative update algorithm on the latent factors to minimize the Kullback–Leibler divergence of the Poisson distributions in the original training tensor versus those represented in its decomposition. Consider a tensor of dimension $D$ and shape $N_1, \dots , N_D$. Each element of this tensor can be thought of as an independent realization from a Poisson distribution. The rate 
\[
\lambda_{i_1, \dots, i_D} \quad , (1 \leq i_1 \leq N_1, . . . , 1 \leq i_D \leq N_D)
\]
is derived from a rank $R$ CPD as:

\begin{equation}
\mathcal{X}_{i_1, \dots, i_D} \sim \text{Poisson}
 \: (\lambda_{i_1, \dots, i_D})
\end{equation}
\begin{equation}
\label{eq:reconst_lamb}
\lambda_{i_1, \dots, i_D} = \sum_{r=1}^{R} \gamma \,r \prod_{d=1}^{D} \theta^{d}_{r,i_d}
\end{equation}
Here, $\theta^{d}_{r}$ represents the \textit{d}-th dimension (or factor) in the \textit{r}-th component.

When training this model, the multiplicative update algorithm tweaks the latent factors to maximize the joint log-likelihood of observed values. The log-likelihood function enables fast training on sparse tensors, as it computes only non-zero values, making it efficient for naturally sparse SCADA system data.

\subsection{Rank Selection}
\label{MF}

Determining the best rank $R$ for a given application is crucial when seeking concise latent tensor representations. A flawless reconstruction of the input tensor using tensor factorization will capture all of the noise in the data and will not effectively convey information about peer groups or inherent structures (\textit{over-fitting)}. Conversely, a rank that is too low might omit critical patterns (\textit{under-fitting)}. 

We chose to hold-out on some of our simulated anomalous data as a validation set. We compared area under the curve (AUC) of precision recall (PR) curves for each rank on this validation set to find the rank that best predicts future SCADA messages. We fit the tensor factorization model on the training set on ranks from 1 to 100 (with a step size of 5 between 50 and 100). The rank with the highest PR AUC was chosen as the rank to decompose our tensors. 

\subsection{Poisson Rate Smoothing}
 Due to the sparsity of our tensors, many of the calculated factors are populated with majority zeros. These zero values cause the estimated Poisson rates to be zero during evaluation due to numerical underflow. To address this, we have incorporated two regularizations in our estimation of the training tensor following the original work \cite{Eren2021pyCPAPR}. First, we inflated the counts of our binary tensor so that the tensors average value in the tensor is close to 1. Second, we used a fusion technique, where the factors of a rank-1 decomposition are fused with those of a rank-$R$ decomposition. Given that any column or row in our tensor has at least one non-zero value, we can be confident that our rank-1 factorization will have non-zero factors. Therefore, by fusing a rank-1 tensor with a rank-$R$ tensor we can ensure that every estimated Poisson rate is greater than zero. Below is a representation of our regularization of the estimated Poisson rate $\lambda_{i_1, \dots, i_D}$:

\begin{equation}
\lambda_{i_1, \dots, i_D} = 0.1 \cdot \lambda^{1}_{i_1, \dots, i_D} + 0.9 \cdot \lambda^{R}_{i_1, \dots, i_D}
\end{equation}

\subsection{Anomalous Message Scoring}

Messages are scored based on the Poisson distribution defined by the counts of events observed during training. A P-value can be calculated by the probability of a value occurring at least as large as the value tested under its Poisson distribution $P(X_{i_1},\dots,i_D \geq x \: | \: \lambda_{i_1},\dots,i_D)$, where $\: \lambda_{i_1},\dots,i_D$ calculated as shown in Equation \ref{eq:reconst_lamb} and $X_{i_1},\dots,i_D$ is the test point. Such that, scoring an event is a relatively cheap computation of factor multiplication/addition and can score individual messages in real-time during testing.  Here the obtained p-value will represent the anomaly scores, where the lower p-value will be an indicator for an anomaly.


\section{LANL Electrical Grid Dataset}

We analyzed SCADA messages collected from an electrical grid operated by Los Alamos National Laboratory that provides transmission and distribution service in partnership with Los Alamos County. This effort leverages SCADA messaging from January \& February of 2020 between the SCADA system and geographically distributed RTUs connected via serial channels. Initial training was done using data from the month of January, while simulations used for evaluation were generated based on data from February. 

The observed SCADA messages included a variety of message types but were dominated by messages requesting analog and digital measurements from devices in a given substation. We opted to test our models on analog scan messages due to the abundance of data and the potential for hiding anomalous messages.

An analog scan request is sent from the SCADA controller to each RTU at a high frequency. Since different RTUs often have different configurations and connected devices, responses can have many different subfields which differ between RTUs. Because of this, we focused on analog scan request values that were present in all requests. These variables were then represented as dimensions in our tensors:
\begin{itemize}
    \item \textit{RTU ID}, the address used to communicate with an RTU.
    \item \textit{Number of Points Requested}, the number of analog measurements requested from a given RTU.
    \item \textit{Channel}, an identifier for the serial channel being used to reach the RTU.
    \item \textit{Time}, elapsed time since the last analog scan message was sent to a given RTU.
\end{itemize}

Each RTU in the system is configured with a user-defined number of points which map to analog inputs for the system. These points are typically queried in sequential blocks, but the number of points and indices of these blocks may vary depending on the SCADA system configuration (e.g. a block of 4 high-priority points may be polled at a higher frequency than 12 lower priority points on the same RTU). Our training data included 24 unique RTU IDs, 22 unique number of points requested values, and nine unique channels. These unique values made up the sizes of each dimension in our tensors. The number of unique time bins is dependent on the granularity chosen. We opted to vary the sizes of the time bins depending on the density of messages in a given periodicity. Denser periodicity's were broken up with higher granularity binning, while the time bins of scarce periodicity's were slightly larger. Our tensors that included time (IPT and IPCT) had 4,283 unique time bins. 

\begin{table*}
\caption{TENSOR DETAILS AND "GREY BOX 1" TEST SET P-VALUE STATISTICS FOR RED TEAM AND BENIGN EVENTS. DIMENSION SIZES REPRESENT THE NUMBER OF UNIQUE VALUES FOR EACH VARIABLE. TIME DIMENSION SIZE REPRESENTS THE NUMBER OF TIME BINS. TIME BINS VARY IN SIZE. TENSOR DIMENSION NAME ABBREVIATIONS ARE \textbf{I}: RTU ID, \textbf{P}: NUMBER OF POINTS REQUESTED, \textbf{T}: TIME BINS, \textbf{C}: CHANNEL}
\resizebox{\textwidth}{!}{
\centering
\label{table:1}
\begin{tabular}{@{}lcccccccccc@{}}
\toprule

& \multicolumn{3}{c}{\textbf{Tensor Details}} & \multicolumn{2}{c}{\textbf{Red Team P-value}} & \multicolumn{2}{c}{\textbf{Benign P-value}} \\ 

\cmidrule(l){2-4} \cmidrule(l){5-6} \cmidrule(l){7-8}

Tensor/Matrix Name & Dimension Size & \% Non-Zero & Decomposed Rank & Mean & Std &  Mean & Std  \\
\midrule
IPT (\textbf{ours}) & 24 (\textbf{I}) x 22 (\textbf{P}) x 4,283 (\textbf{T}) & 1.8 & 5 & .184 & .289 & .802 & .343\\
\hline
IPCT (\textbf{ours}) & 24 (\textbf{I}) x 22 (\textbf{P}) x 9 (\textbf{C}) x 4,283 (\textbf{T}) & 0.2 & 5 & .061 & .164 & .731 & .320 \\
\hline
IPC (\textbf{ours}) & 24 (\textbf{I}) x 22 (\textbf{P}) x 9  (\textbf{C}) & 0.5 & 47 & .245 & .221  & 1 & 0 \\
\hline
NMF: I x P & 24 (\textbf{I}) x 22 (\textbf{P}) & 4.7 & 24 & .572 & .456  & .999 & 2.04 x $10^-5$ \\
\hline
NMF: I x C & 24 (\textbf{I}) x 9 (\textbf{P})  & 11.1 & 14 & .413 & .442  & .999 & 9.56 x $10^-6$ \\

\bottomrule
\end{tabular}
}
\end{table*}

\subsection{Simulated Anomalies}
Test sets were generated by an in house simulator previously developed to model the communication between RTUs using a proprietary serial protocol. The simulator replays recorded traffic and introduces additional anomalous messages with randomized values as described below. Each simulation run included roughly 130,000 messages, with around 1000 of those being anomalous. These test sets are meant to simulate an adversary scanning the network to identify RTUs and their configurations (normal message values). In order to do so, the adversary queries different RTUs, number of points requested, and channels based on their limited knowledge of the SCADA system. Reconnaissance is often necessary in order to acquire the necessary information required to attack or degrade the system. The test sets used represent progressively more sophisticated adversaries which attempt to circumvent detection during the reconnaissance phase by blending in with typical system traffic.

\paragraph{Black Box Attack}
This simulation represents an adversary with minimal knowledge about the system that tries to enumerate values for RTU and channel while also trying to learn configuration values such as the number of points configured on the RTU. Each variable's value is chosen randomly from the range of possible values defined by the protocol, with the exception of channel which is randomly selected from the set of valid channels. Since the range of possible values is much larger than the actual number of unique values for each dimension, the vast majority of the adversary's messages contain values that are not present in the training data, making them relatively simple to detect.  

\paragraph{Grey Box Attack 1}
Grey Box attack 1 represents an adversary that has some knowledge of the system (e.g. the expected values for RTU ID, Number of Points Requested, and Channel) but without knowing exactly which combinations of these attributes are normal for the system. The anomalies in this test set are closer to normal system behavior than those in the Black Box Attack and therefore harder to detect.

\paragraph{Grey Box Attack 2}
The second Grey Box Attack is similar to the first one in that the adversary has foreknowledge of likely values, but this time the adversary knows which channel and RTU pairs the system uses but not how many measurements the RTU is configured for. The anomalies in this test set are the most similar to normal system behavior and therefore the hardest to detect. 

Table \ref{table:2} shows the number of total messages and anomalous messages of the three test sets, which were simulated based on real messages sent from February 1st to 5th, 2020. 

\begin{table}[h]
\caption{Characteristics of Simulated Test Sets Depicting Different Attack Scenarios: Breakdown of Total Messages and Number of Anomalous Messages Per Test Set}

\resizebox{\columnwidth}{!}{
\centering
\label{table:2}
\begin{tabular}{c|c|c}
\toprule
\textbf{Test Set} & \textbf{Total Messages} & \textbf{Anomalies} \\
\midrule
Black Box Attack & 130,029 & 1,009  \\
Grey Box Attack 1 & 129,873 & 1,000 \\
Grey Box Attack 2 & 129,871 & 1,000 \\
\bottomrule
\end{tabular}
}
\end{table}

\subsection{Tensor Construction}
We constructed three tensors: a binary tensor and two count-value tensors. The binary tensor omits the time dimension, allowing us to assess time's influence on results.

A common way of representing time in a tensor is by binning it into minute of the hour, hour of the day, day of the week etc. This did not work for SCADA messages because of their high frequency and shifting time of occurrence. Furthermore, minor changes in the system over time would alter this shift. When representing time this way, the tensor's sparsity hit 100\%, with ROC and PR curves resembling random chance. We address this issue by representing time as the number of milliseconds between consecutive messages sent to the same RTU ($\Delta$ Time). 

The dimensions of the three tensors that we constructed for CP-APR are: \textit{RTU ID - Number of Points Requested - $\Delta$ Time} (IPT), \textit{RTU ID - Number of Points Requested - Channel - $\Delta$ Time} (IPCT), and \textit{RTU ID - Number of Points Requested - Channel} (IPC). IPT and IPCT used count values as the tensor entries, while the other tensors used inflated binary entries. All values have been binned to anonymize the specific values and time bins are not homogeneous. A high degree of variability is observed along the time dimension, making it difficult to detect anomalies purely based on time. With the IPC tensor, we found that using binary values worked a bit better. The only values in a binary tensor are zero and one. For the binary tensor IPC, a count of one indicates that combination of RTU ID, Number of Points Requested, and Channel occurred at least once. For the IPT and IPCT tensors, the count value indicates the number of times that that combination of values occurred in January 2020. Each of the tensors (IPT, IPC, and IPCT) were factorized using CP-APR (specifically, pyCP-APR \cite{Eren2021pyCPAPR}). 

SCADA messages result in immensely sparse tensors, where only a small fraction of the tensor is made up of non-zero elements. Sparsity is the percentage of entries in the tensor that are non-zero. Our IPT tensor had a sparsity of 1.8 percent. Zero values that comprise the majority of the tensor do not need to be stored in memory, allowing us to deviate from dense tensor representation. This fact is part of the reason why sparse tensor decomposition is efficient. Instead, sparse tensors can be stored as a list of coordinates with a corresponding list of non-zero values (\textit{COO} format). Table \ref{table:1} shows statistics for each tensor.

\section{Experiments and Results}
We trained our models on real world data which we assumed to be 100\% benign. We then simulated normal messages over a different time period with different anomalous messages sprinkled in. An example of an anomalous message would be a message to an RTU with a different number of points requested than what it normally receives. Our models are intended to be incorporated as a continuous monitoring tool, where new messages are scored against an existing model, and periodic batch retraining is scheduled to enhance the model's accuracy as the grid changes over time.

We identify anomalous messages by calculating a P-value for each element within the observed tensor. Model performance is assessed numerically using the area under the curve (AUC) of the the receiver operating characteristic (ROC) and Precision Recall (PR) curves. The AUCs of these two curves describe how effectively the model gives anomalous messages lower P-values compared to benign ones. The ROC curve plots the true positive rate (TPR or recall) over the false positive rate (FPR, percentage of benign messages that were false positives), while the PR curve plots the precision (percentage of positive results that were actually anomalous) over the recall (percentage of anomalies that were detected). 

\begin{figure}[h]
    \centering
    \includegraphics[width=1\linewidth]{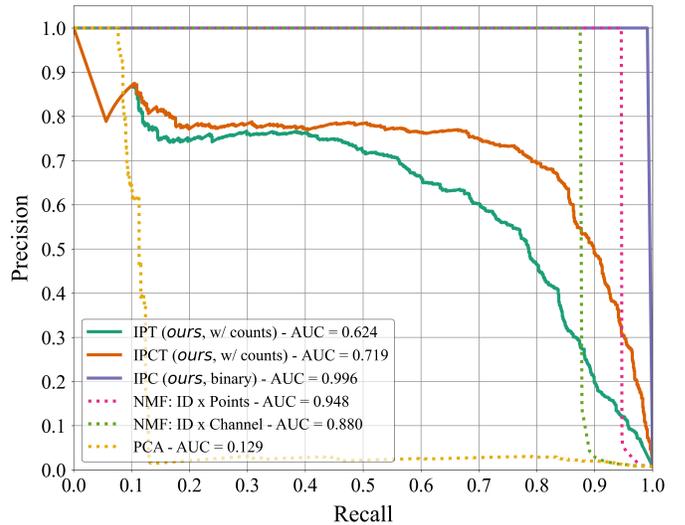}
    \caption{Grey Box 1 PR Curves. PR curve is a better indicator of accuracy in our tests because the proportion of anomalous vs benign messages is very low. Here we see that the tensors with the time dimension performed worse than the models that did not include time.}
    \label{fig:pr}
\end{figure}

When a PR curve is flat at precision = 1 and then shows a steep drop at recall = .99, like the purple IPC curve in Figure \ref{fig:pr}, it means that up to a certain P-value threshold every message with a P-value below that cutoff was in fact anomalous and that those anomalies detected were 99\% of the total number of anomalies in the test set. The steep drop means that in order to detect the next true positive, the P-value threshold had to be raised to a point where a high number of benign messages were also classified as anomalous (false positives). The IPC tensor's near perfect PR AUC of .99 tells us that 99\% of the anomalies in the test set were abnormal on at least one dimension other than time. This could also partly explain why the TD and NMF models that did not consider timing performed so much better than the three models that did include time. While the existing anomalies predominantly exhibited irregularities in factors other than the temporal dimension, introducing anomalies that maintain correct values but occur at abnormal times might make a more challenging test for the models. Such anomalies would help to evaluate the depth and capability of timing-aware models in comparison to those that do not consider timing. If these anomalies were to occur in the real world it would imply that the adversary has somehow gained complete knowledge of normal system behavior. 

While table \ref{table:3} includes ROC AUC scores, it is important to note that these scores are quite high due to the disparity between the number of anomalies and benign messages. With a ratio of roughly 1,000 anomalies to 130,000 benign messages, the vast majority of the data was non-anomalous, which could skew the perceived performance. For this reason, PR AUC scores are better indicators of accuracy in our tests.  

\begin{table}[h]
\caption{AUC of ROC and PR Scores for Each Model. Best PR Score for Each Attack is Highlighted.}
\resizebox{\columnwidth}{!}{
\centering
\label{table:3}
\begin{tabular}{@{}c|c|c|c|c|c|c}
\toprule
& \multicolumn{2}{c}{\textbf{Black Box}} & \multicolumn{2}{c}{\textbf{Grey Box 1}} & \multicolumn{2}{c}{\textbf{Grey Box 2}} \\
\cmidrule(l){2-3}\cmidrule(l){4-5}\cmidrule(l){6-7} 

\textbf{Model} & ROC & PR & ROC & PR & ROC & PR\\

\midrule
IPT (\textbf{ours}) & .999 & .970  & .983 & .624 & .993 & .746\\
IPCT (\textbf{ours}) & .999 & .982  & .995 & .719 & .995 & .798\\
IPC (\textbf{ours}) & .999 & \textbf{.999}  & .996 & \textbf{.996} & .973 & \textbf{.973}\\
NMF: IxP & .997 & .996  & .970 & .948 & .969 & .947\\
NMF: IxC & .984 & .958 & .936 & .880 & .651 & .105\\
PCA & .939 & .572  & .819 & .129 & .877 & .177\\
\bottomrule
\end{tabular}
}
\end{table} 
Our tests reveal that our tensor decomposition model is capable of pinpointing anomalies across various modalities. Moreover, incorporating an additional dimension of \textit{Channel} into the analysis enhances the profiling of intricate activities. Table \ref{table:3} shows the AUCs of our three tensor decomposition models IPT, IPCT and IPC, along with the models that we used as baseline comparisons. We expected IPCT to outperform IPT on the first two tests, when many anomalies have incorrectly mapped channels, but an interesting takeaway from comparing the results of the two tensors is that IPCT also had higher ROC and PR AUCs on the test where every single anomaly had a correctly mapped channel value for its RTU. This indicates that increasing dimensions generally improves anomaly detection. The same theme can be found when comparing NMF: IxP to IPC, which is effectively a higher order version of NMF: IxP with channel as an added dimension.  

The PCA model included the same four variables as the IPCT tensor, and yet showed a dramatic decrease in AUCs on all tests. This demonstrates that reducing dimensionality of the same variables loses much of the intricate details hidden in the data. 

These results show that the binary tensor IPC, which does not include time, generated the highest AUCs. This implies either that the distribution of messages across the time component is too varying to add value or that further work is needed to detect anomalies on the time dimension. Furthermore, testing with a higher number of messages that are only anomalous on the time dimension could be a way to evaluate more robust models that can not only detect scanning, but also messages sent by adversaries with complete knowledge of normal system behavior. 

\section{Conclusion}

We present a novel application of tensor decomposition as a multidimensional anomaly detection method for SCADA messages. Our results demonstrate that higher-dimensional representations improve anomaly detection. This is attributed to tensor decomposition's capability to derive more detailed SCADA system behavior, characterizing events across several dimensions concurrently. 

However, the inclusion of time, despite experimenting with varying granularity and representation, did not enhance the accuracy of our models. This hints at inherent complexities in the temporal aspects of SCADA messages, warranting different or additional techniques. \cite{doi:10.1137/1.9781611977653.ch10} and \cite{DBLP:journals/corr/abs-1912-09009} have recently introduced methods for estimating optimal granularity—a topic for future exploration including temporal tensor granularity estimation, which may be beneficial to explore in the future work. Additionally, the fact that our simulated scanning behavior was easy to detect allowed some of our models to score near perfect. We would like to explore testing on anomalies that are even more similar to normal behavior and therefore harder to detect. Such anomalies would likely occur across different message types. 

Tensor decomposition outperforms PCA and NMF at detecting anomalies similar to normal SCADA behavior. Furthermore, the fact that adding an extra dimension of Channel increased accuracy on every test, including one where zero of the anomalies had an incorrect channel, indicates that a tensor decomposition with even higher dimensions may yield further improvements. With the increasing complexity and scale of electrical grids\cite{YADAV2021100433}, the demand for sophisticated anomaly detection techniques is evident, emphasizing the need for methods that can fully capture the complexities of electrical grid systems without the need for labels. In future work we intend to explore incorporating more message types and values into a higher order tensor and to explore other potential approaches to better address the temporal aspects of SCADA messages.

\section*{Acknowledgment}
This manuscript has been assigned LA-UR-23-30012 Version 3. Research presented in this paper was supported by the Information Science and Technology Institute's Cybersecurity Science Research Program, and by the Laboratory Directed Research and Development program of Los Alamos National Laboratory (LANL) under project numbers 20190020DR and 20200666DI. Data used in this effort was obtained through a collaboration with LANL's Utilities and Infrastructure division. LANL is operated by Triad National Security, LLC, for the National Nuclear Security Administration of the U.S. Department of Energy (Contract No. 89233218CNA000001).

\bibliographystyle{IEEEtran}
\bibliography{references}

\end{document}